# *Women, Infamous, and Exotic Beings*: A Comparative Study of Honorific Usages in Wikipedia and LLMs for Bengali and Hindi

**Sourabrata Mukherjee**[1,2] [*] [†], **Atharva Mehta**[1] [*], **Sougata Saha**[1]
**Akhil Arora**[3], **Monojit Choudhury**[1]

[1]Mohamed bin Zayed University of Artificial Intelligence,
[2]Microsoft Research India, [3]Aarhus University
[1]{atharva.mehta, sougata.saha, monojit.choudhury}@mbzuai.ac.ae
[2]t-somukherje@microsoft.com, [3]akhil.arora@cs.au.dk

## Abstract

The obligatory use of third-person honorifics is a distinctive feature of several South Asian languages, encoding nuanced socio-pragmatic cues such as power, age, gender, fame, and social distance. In this work, (i) We present the first large-scale study of third-person honorific pronoun and verb usage across 10,000 Hindi and Bengali Wikipedia articles with annotations linked to key socio-demographic attributes of the subjects, including gender, age group, fame, and cultural origin. (ii) Our analysis uncovers systematic intra-language regularities but notable cross-linguistic differences: honorifics are more prevalent in Bengali than in Hindi, while non-honorifics dominate while referring to infamous, juvenile, and culturally "exotic" entities. Notably, in both languages, and more prominently in Hindi, men are more frequently addressed with honorifics than women. (iii) To examine whether large language models (LLMs) internalize similar socio-pragmatic norms, we probe six LLMs using controlled generation and translation tasks over 1,000 culturally balanced entities. We find that LLMs diverge from Wikipedia usage, exhibiting alternative preferences in honorific selection across tasks, languages, and socio-demographic attributes. These discrepancies highlight gaps in the socio-cultural alignment of LLMs and open new directions for studying how LLMs acquire, adapt, or distort social-linguistic norms. Our code and data are publicly available at https://github.com/souro/honorific-wiki-llm

## 1 Introduction

Honorifics are powerful linguistic tools that encode social hierarchies and cultural norms (Agha, 1998). While English relies on titles (e.g., Mr., Sir) without grammaticalized honorifics (Brown, 1987), many South Asian languages (Dryer and Haspelmath, 2013; Helmbrecht, 2013)[1], including Hindi and Bengali[2], require speakers to explicitly mark formality in pronouns and verb forms (Ferschke et al., 2013). For example, a simple sentence like "*She is a doctor*" in Hindi must commit to either an **honorific** ("*ve* chikitsak *hain*") or **non-honorific** ("*vaha* chikitsak *hai*") construction. Such choices in Hindi and Bengali are not optional but essential, signaling respect, age, status, and familiarity.

Honorific usage is deeply rooted in cultural norms, highly context-sensitive, and varies significantly across age groups, regions, and genres (Friedrich, 1972; Agha, 2007; Ferschke et al., 2013). Improper use of honorifics, for example, omitting them when referring to respected figures in Hindi or Bengali, can be perceived as disrespectful, potentially leading to public backlash or reputational harm, especially in formal or public discourse contexts such as LLM chat, writing assistance, or educational platforms. Given that most large language models (LLMs) are trained on vast datasets, including Wikipedia, understanding how honorifics are represented on Wikipedia and how LLMs internalize these norms is of paramount importance. However, large-scale empirical studies on honorific usage remain scarce, with most research being limited in scale and focused on specific languages (Brown et al., 1960; Harada, 1976; Agha, 1998; Bhatt, 2012). To date, there has been no systematic study examining how LLMs use honorifics across languages or cultural contexts.

To address this gap, we conduct a large-scale analysis of 10,000 Wikipedia articles each in Hindi (HI) and Bengali (BN) [3], sampled across eight socio-demographic dimensions, to examine patterns of honorific usage. The analysis employs GPT-4o as the annotation tool, which exhibits strong performance as confirmed by human evaluations (see Section 3). This study is particularly significant because, although Wikipedia is a formal platform, its editorial guidelines[4] do not explicitly define when or how to use honorifics in languages where such forms are grammatically obligatory. This lack of explicit guidelines can lead to the appearance of implicit and extant sociocultural biases in honorific usage.

---

[*]Equal contribution
[†]Work done while interning at MBZUAI

[1]https://en.wikipedia.org/wiki/Honorifics_(linguistics)
[2]Linguistic background on honorific usage in Hindi and Bengali is detailed in Appendix B.
[3]The rationale behind language selection and Wikipedia as a data-source are provided in Appendix A and C respectively.

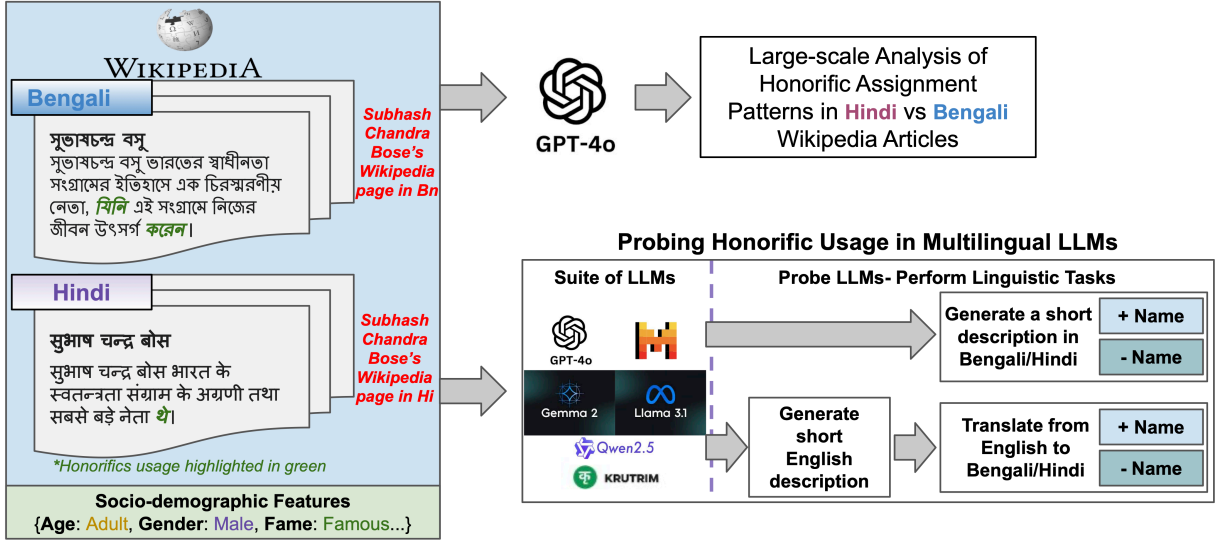

Figure 1: Illustration of our end-to-end analysis pipeline.

Moreover, as LLMs rely heavily on such online sources for training, it is unclear whether they accurately reflect the honorific usage patterns observed in Wikipedia, especially across diverse socio-demographic contexts. To investigate this, we evaluate six state-of-the-art LLMs through controlled generation and translation tasks involving 1,000 culturally balanced entities, sampled using the same socio-demographic dimensions and formality settings (Section 4). Our key contributions are:

i. A first-of-its-kind large-scale empirical study of honorific usage across 10,000 HI and BN Wikipedia articles, analyzing cross-lingual differences across eight diverse socio-demographic dimensions.

ii. A framework for evaluating honorific behavior in multilingual LLMs, using generation and translation tasks to measure their usage patterns across similar socio-demographic features and formality settings as Wikipedia.

iii. The public release of a dataset comprising 20,000 annotated Wikipedia articles and LLM outputs for 1,000 entities.

The rest of the paper is organized as follows: Section 2 reviews related work. Section 3 details our Wikipedia data annotation and analysis. Section 4 presents the LLM probing setup and comparison with Wikipedia patterns. Section 5 concludes with key insights and open questions.

## 2 Related Work

The study of honorifics has traditionally been situated within the fields of theoretical linguistics, sociolinguistics, and pragmatics. Early foundational work, such as by Brown and Gilman (1968), proposed that honorific usage is governed by the competing forces of power and solidarity within social relationships. Similarly, Agha (1994) provided a comprehensive account of honorific registers, emphasizing how language mediates social stratification and cultural norms. These theoretical frameworks have significantly shaped our understanding of honorifics, yet they remain largely qualitative and tailored to specific languages or cultures. While they provide crucial insights and starting points for identifying relevant features, they are not sufficient for evaluating or building text generation models (e.g., LLMs). Moreover, the absence of a universal theory of honorific usage implies that each language or cultural setting requires its own quantitative exploration. Other works, such as Wales (1983) and Friedrich (1972), argued for a broader interactional perspective, suggesting that pronoun variability, and by extension, honorific usage, depends not only on hierarchical relationships but also on dynamic contextual factors such as familiarity, emotional stance, and discourse setting.

---
[4]Wikipedia:Editing Policy

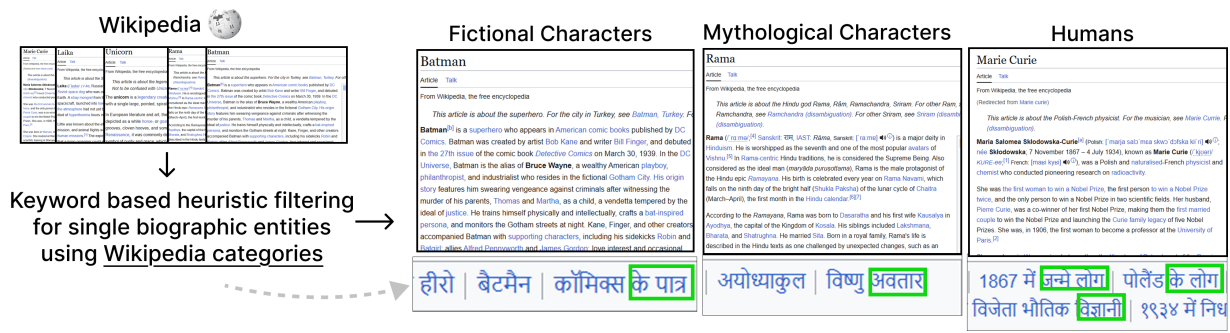

Figure 2: We developed a pipeline for extracting entities from Wikipedia. In the initial stage, we obtained all the entities for Hindi and Bengali from Wikidata. To identify single-entity articles, we relied on Wikidata categories and further applied keyword-based filtering over article categories (e.g., में जन्मे "born in" for humans, देवता for deities, आतंकवादी for infamous figures, etc.). From these, we performed random sampling across different keywords and entity types to construct a balanced final corpus of 10,000 entities each for Hindi and Bengali. To ensure accuracy, both the authors and human annotators manually validated a random subset of the extracted pages.

In the context of Hindi & Bengali, several studies have explored honorific phenomena. Bhatt (2012, 2015) examine honorifics in Hindi, emphasizing their complexity in social relationships but noting the absence of clear rules guiding usage, often relying on instinct, and therefore, making it challenging to learn by non-native speakers. Hakim (2014) investigated American students' difficulties in using Bengali honorifics, despite strong grammatical proficiency. Uddin (2019) compared second-person pronoun usage in Bengali and English, finding that Bengali's system is more intricate, yet their study was largely based on intuitive observations rather than empirical data. Bhattacharya et al. (2005) explored a rule-based method for generating honorific forms in Bengali. Chatterji et al. (2012) studied that, pronominal forms are difficult to translate from English, which lacks distinct honorifics, to BN or HI, but they also examine a small set of pronouns and miss out the broader contextual influences.

Overall, these studies provide valuable insights into honorific usage, but a large-scale, data-driven, and computational understanding is still missing; our work explores to take an initial step toward filling this gap.

## 3 Honorific Usage Patterns in Wikipedia

Wikipedia promotes a formal and encyclopedic tone with editorial guidelines that are largely consistent across languages.[5] Given this editorial formality, we hypothesize: **The third-person honorific usage in Wikipedia articles should exhibit consistency across languages.** To test this, we extract and annotate a large corpus of Hindi and Bengali Wikipedia articles (see Section 3.1) and conduct a systematic analysis to evaluate the hypothesis (see Section 3.1).

### 3.1 Annotation Methodology

**Wikipedia Data Extraction and Sampling.** Using the *Wikidata SPARQL API*[6], we narrow down an initial pool of 10 million articles to the set of all articles in Hindi and Bengali to 28,755 and 51,293 pages, respectively. Finally, we extract the introductory text for each entity, which further reduces the pool to 24,505 for Hindi and 43,108 for Bengali.

From this curated pool, we extract all pages that correspond to biographical single entities. A single-entity article is defined as a Wikipedia page dedicated to one individual, such as a human (eg: Narendra Modi), animal (eg: Paul the Octopus), deity (eg: Lord Rama), or fictional character (eg: Batman). To construct clean and focused datasets, we first identify and filter all single-entity articles and their corresponding URLs for Hindi and Bengali using heuristic methods. Specifically, we rely on Wikidata categories and supplement them with keyword-based filtering of article categories (e.g.,

---
[5]See Help:Editing and Wikipedia:Editing policy
[6]https://query.wikidata.org/

में जन्मे "born in" for humans, देवता for deities, आतंकवादी for infamous figures). For illustrative examples and an overview of the heuristic pipeline, refer to Figure 2. From these, we performed random sampling across different keywords and entity types to construct a balanced final corpus of 10,000 entities each for Hindi and Bengali. To ensure accuracy, both the authors and human annotators manually validated a random subset of the extracted pages. Detailed data statistics are provided in Table 3 in Appendix D.

**Socio-Cultural Determinants of Honorific Usage.** To investigate how honorific usage in Wikipedia articles reflects deeper socio-cultural dynamics, we define a set of eight socio-demographic features (see Table 1) grounded in the principles of various cultural and sociolinguistic theories. These features were chosen based on their established relevance in shaping linguistic choices tied to politeness, social hierarchy, and identity construction. For instance, theories of linguistic politeness and honorification (Brown, 1987; Irvine, 1996) emphasize how factors like age, gender, and status directly influence the degree of formality or deference encoded in language. Similarly, work on indexicality in language (Silverstein, 2003; Eckert, 2012) demonstrates how the invocation of categories such as cultural nativeness or mythical status functions symbolically to position subjects along culturally meaningful axes of respect and familiarity. Moreover, these categories align with known triggers of honorific register-shifting in South Asian languages (Agha, 2007), making them robust predictors for the observed linguistic patterns in our study.

**LLM-based Automatic Annotation.** To enable large-scale analysis across 10 000 Wikipedia articles per language, we employed LLMs to automatically annotate socio-demographic features (see Table 1) and identify third-person pronouns and verb forms (see prompt details in Appendix I), focusing on the first paragraph of each article to reduce noise. Among the models tested for automatic annotation, *GPT-4o* achieved the highest agreement with human annotations indicated by Cohen's kappa coefficient and percentage agreement: $\kappa = 0.47$ (BN: 85%) and $\kappa = 0.78$ (HI: 93%) validated by native speakers on 100 randomly selected articles per language (details in Appendices G and K). The agreement between the human annotators was even greater: BN: $\kappa = 0.89$ (98%) and HI:

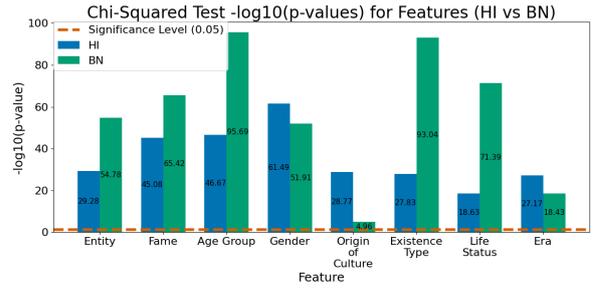

Figure 3: Results of the chi-square significance test evaluating the association between socio-demographic features and honorific usage.

$\kappa = 0.97$ (99%), establishing the reliability of the GPT-4o annotations. In comparison, GPT-3.5 produced lower agreement ($\kappa = 0.13$/BN, $0.19$/HI), and LLaMA3.1-8B performed poorly with negative scores ($\kappa = -0.13$/BN, $-0.12$/HI). Detailed evaluation metrics are provided in Table 4 and visualizations of human agreement in Figures 7 and 8, both in Appendix E. The statistics of the annotated dataset are summarized in Table 3 Appendix D).

### 3.2 Analysis

Overall, 86.6% of Bengali articles use honorific pronouns and/or verb forms, as compared to 77.9% in Hindi.

Since the distribution of entity types (see Table 3 in Appendix D) in our dataset do not show any language specific bias, this observation must then be attributed to the natural tendency or norm for BN speakers to use honorifics more often than HI speakers while refering to an entity in third person in a formal/broadcasting setting. **Indicating that our initial hypothesis is wrong:** language-specific variations in honorifics usage do exist in Wikipedia.

Furthermore, to examine the influence of socio-demographic features on honorific usage across languages, we conducted a chi-squared significance test (Pearson, 1900). Specifically, we tested whether the choice of honorific or non-honorific forms is significantly associated with the socio-demographic features. The results for Bengali (BN) and Hindi (HI) are shown in Figure 3, with the complete chi-squared heatmap provided in Figure 9 with additional details in Appendix F. The analysis revealed that all features are statistically significant (p < 0.05), indicating that our considered socio-demographic features play a important role in shaping honorific usage patterns in Wikipedia articles.

| Feature | Values | Explanation |
| --- | --- | --- |
| Entity Type | God, Human, Animal, Other Being | Distinguishes deities (worshipped divine figures), humans, animals (often symbolic or famous), and other beings (for example tree:The Great Banyan). Useful as honorific norms differ sharply across these types. |
| Fame | Famous, Infamous, Controversial | Famous: widely admired figures; Infamous: entities with unambiguously negative reputation (e.g., R. E. Dyer); Controversial: figures with polarized perception (e.g., Indira Gandhi). This distinction aligns with pragmatics literature and was validated via human annotators. |
| Age Group | Juvenile, Adult, Old, Not Applicable | Juvenile: entities younger than 18; Adult: between 18 and 60; Old: above 60. N/A applies where age is not meaningful (e.g., deities, fictional beings). |
| Gender | Male, Female, Gender Neutral/Non-Specific | Classifies perceived gender. Gender neutrality applies where the subject is non-gendered (e.g., abstract beings, certain animals) or explicitly non-binary. Gender distinctions allow us to capture biases in honorific attribution. |
| Origin of Culture | Native, Exotic | Native: from the same culture as the language (e.g., Tagore in Bengali); Exotic: from outside that culture (e.g., Shakespeare in Bengali). This distinction helps capture how "insider" vs. "outsider" status shapes honorific use. |
| Existence Type | Real, Fictional, Mythological | Real: historical or living persons/animals; Fictional: literary or cinematic characters; Mythological: deities and legendary beings. The category reflects how cultural grounding influences linguistic treatment. |
| Life Status | Alive, Dead, Not Applicable | Tracks whether the entity is living or deceased at the time of writing. Honorific use can shift after death (e.g., increased reverence). N/A applies to beings beyond life/death (e.g., gods, mythological entities). |
| Era | Historical, Modern | Threshold of 1800 reflects socio-linguistic shifts around the colonial onset in South Asia (1757 Plassey, 1857 Sepoy Mutiny; (Ray et al., 1966)). Historical: pre-1800 figures; Modern: post-1800. This allows us to test whether historical vs. modern figures are treated differently in terms of honorific usage. |

Table 1: Socio-demographic features used in our study, their values, and compact explanations. For detailed annotation guidelines, see Appendix G.

To identify the importance of each feature, we modeled the data using a logistic regression function and analyze the coefficients of each feature to predict the honorific usage patterns. Figure 4 shows the regression coefficient for various values of these features (for a detailed correlation analysis of all features with respect to their honorific and non-honorific usage, refer to Figure 12 in Appendix M.).

Our findings are summarized as follows. In terms of **Fame**, famous individuals are addressed with honorifics in both languages, with coefficients of BN: $0.90$ and HI: $0.79$. This aligns with the cultural norm of showing respect to notable figures, as seen with real-world personalities like *Subhash Chandra Bose*, goddess *Durga*, and the epic hero *Arjuna*. Conversely, infamous individuals are consistently referred to without honorifics, as reflected by the negative coefficients of BN: $-0.65$ and HI: $-0.81$. Examples include the demon *Putana* from the epic Ramayana and *R. E. Dyer*, who is associated with the Jallianwala Bagh massacre. The data further suggests that Hindi may be more rigid in withholding honorifics from individuals with negative reputations, reflecting stricter societal boundaries regarding respect.

For the **Age Group** feature, both languages follow a similar pattern, with older individuals receiving honorific treatment and younger individuals typically addressed with non-honorific forms. Bengali shows a much stronger preference for honouring elders, with an honorific to non-honorific ratio of 14.45, compared to 6.51 in Hindi. For instance, *Gangubai Kothewali*, despite her old age, is referred to with non-honorifics in Hindi, whereas honorifics are used to refer to her in Bengali. The coefficient for older individuals in Bengali (0.92) is nearly triple that of Hindi (0.30), suggesting a deeper cultural emphasis on respecting elders in Bengali. For juveniles, both languages display negative coefficients, with BN:−0.61 and HI:−0.84, indicating a stronger association of non-honorific language with youth.

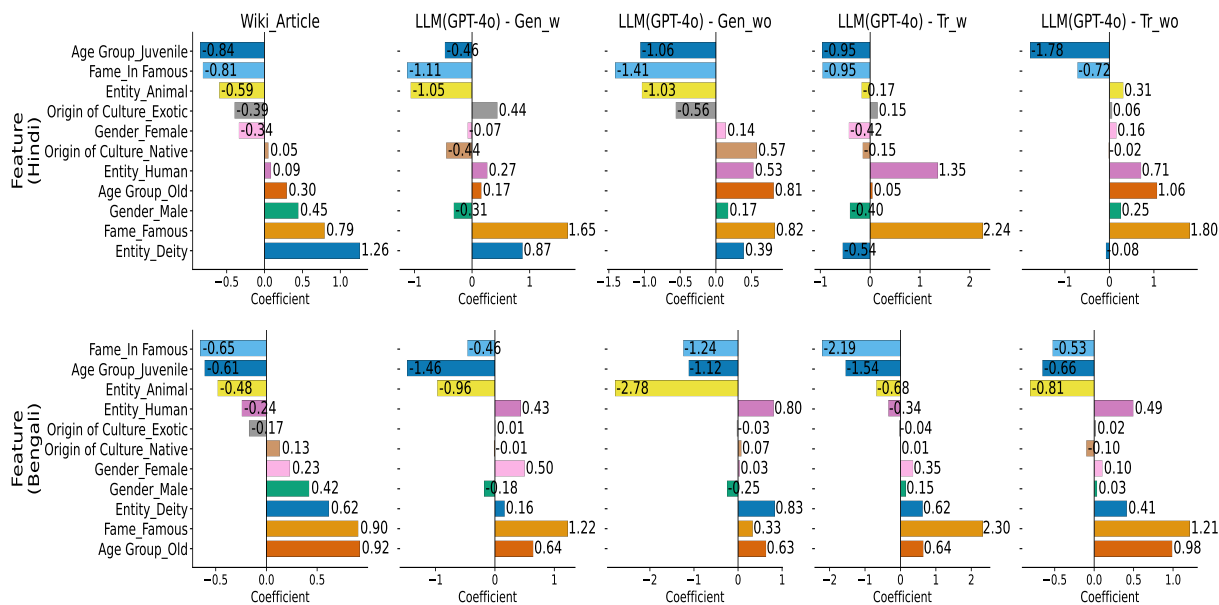

Figure 4: Logistic regression coefficient comparison for Bengali (BN) and Hindi (hi) across various socio-demographic features for GPT-4o across 4 tasks: Gen_w, Gen_wo, Tr_w & Tr_wo (for details about this tasks see Section 4.1) and Wikipedia Articles. Positive values indicate more frequent honorific usage, while negative values suggest non-honorific usage.

Regarding the **Origin of Culture**, both Hindi and Bengali favour honorifics for native individuals, with a coefficient of 0.05 for HI and 0.13 for BN. Figures such as *Nandi*, *Jatayu*, and *Sourav Ganguly* are addressed with honorifics in both languages, reflecting a shared cultural practice of showing respect to native entities. In contrast, exotic individuals are more likely to be referred to with non-honorific forms, as indicated by the negative coefficients (BN:−0.17 and HI:−0.39). This is exemplified by Figure like *Lord Lytton*, a British person, consistently addressed without honorifics in both languages. This demonstrates a common cultural norm across the languages, where the same norms do not apply to entities from within and external to the culture.

Despite these similarities, there are notable differences between the two languages. Bengali consistently exhibits higher honorific usage, especially for elders, famous individuals, and male figures. In contrast, Hindi shows stronger non-honorific tendencies, particularly for infamous and exotic individuals.

Interestingly, **Gender** reveals a sharp contrast between Hindi and Bengali. While male individuals in both languages are frequently addressed with honorifics (BN: 0.42 and HI: 0.45), for females, Bengali shows a smaller than male but nevertheless a positive coefficient of 0.23, whereas in Hindi, the coefficients is negative, −0.34, showing a higher tendency to refer to women with non-honorifics. *Rubina Ali*, and *Malala Yousafzai*, who receive honorifics in Bengali, are referred to with non-honorific pronouns in Hindi. In general, both *Male* (honorific to non-honorific ratio of 7.37) and *Female* (ratio of 5.50) figures in Bengali are more likely to be addressed with honorifics compared to their Hindi counterparts (Male: 4.52, Female: 2.12), suggesting a stronger cultural inclination toward honorifics in Bengali regardless of gender. This highlights a gender bias in Hindi, perhaps reflective of a similar bias in society.

For **Entity Type**, the honorific usage in Wikipedia articles shows clear distinctions across entity types. Deity entities exhibit the strongest positive association with honorific forms, reflected in high coefficient values in both Hindi (+1.26) and

Bengali (+0.62) (for example *Durga, Kali and Lakshmi* in Hindi and *Saraswati, Ram, Narayan and Brahma*). In contrast, Animal entities are consistently associated with non-honorific forms, as indicated by negative coefficients (-0.59 in Hindi, -0.48 in Bengali) (for example *Airavat, Pheonix, Nandini(cow)* in Hindi and Bengali). Human entities show marginal effects, with slightly positive coefficients in Hindi (+0.09) and negative values in Bengali (-0.24), suggesting variability in how honorifics are used based on other cues (i.e., famous or infamous).

Regarding the **Era**, we observe opposing trends across the two languages. For modern-era entities, Hindi shows a higher tendency toward honorific usage (0.23), while Bengali favors non-honorifics (-0.18). Conversely, for historical entities, Hindi leans toward non-honorifics (-0.58), whereas Bengali exhibits a positive association with honorifics (0.18).

For **Life Status**, both languages predominantly use non-honorifics when referring to living individuals (HI: -0.28, BN: -0.34). However, for deceased entities, the patterns diverge: Hindi shows a near-neutral coefficient (-0.02), while Bengali demonstrates a stronger non-honorific tendency (-0.30).

We conclude by stating that **our null hypothesis is rejected**, as the patterns of honorific usage across sociocultural features differ across attributes and languages.

## 4 Probing Honorific Usage in LLMs

Given that Wikipedia comprises a significant portion of many LLMs' training data, we hypothesize that patterns of inconsistent honorific usage and their inconsistencies, present in Wikipedia (see Section 3.2), will be reflected in LLM outputs- i.e, **third-person honorifics usage in LLMs should reflect similar patterns as exhibited in Wikipedia**. To test this, we design a controlled LLM probing experiment (Section 4.1) and provide a comparative analysis of honorific patterns in Wikipedia and LLM-generated content (Section 4.2).

### 4.1 LLM Probing Setup

**LLM Model Details** Our analysis focuses on large language models that advertise multilingual support for Hindi and Bengali, spanning the proprietary GPT-4o(Hurst et al., 2024); general open-source systems such as Llama-3.1-8B-Instruct(Grattafiori et al., 2024), Gemma-7B-Instruct(gem, 2001), Mistral-7B-Instruct-v0.3(Jiang et al., 2023), and Qwen-2.5-7B-Instruct(Team, 2024); and the Indic open-source model Krutrim-2-Instruct(Kallappa et al., 2025). This list of models enable a comparative study of honorific pronoun and verb usage across proprietary, mainstream open-source, and region-focused LLMs.

**Evaluation Task Design** To probe honorific usage, we devise two complementary tasks, each executed in **Hindi** and **Bengali**. For all probing tasks, we use the same 10k Wikipedia data we collected(explained in Section 3.1) for Hindi and Bengali, which is annotated by GPT-4o for sociodemographic features(given in Table 1).

Given the annotated bundle of sociodemographic attributes, the model must write a short paragraph about each entity in the annotated data. We run two variants: with an explicitly supplied name (GEN_W) and without a name (GEN_WO). Comparing these variants reveals whether encountering a potentially "real" name alters honorific choice.

The model first completes the paragraph in English, whose grammar lacks obligatory honorific marking, and then translates *its own* text into the target language. We again test TR_W and TR_WO. Using model-generated English rather than Wikipedia prevents training-data leakage and avoids verbatim reproduction of encyclopedic prose. Together, the four settings (GEN_W, GEN_WO, TR_W, TR_WO) let us contrast honorific behavior when the model generates content from scratch versus when it transfers an existing narrative. The detailed prompts for all tasks are shown in Appendix H.

**Prompt Construction and Annotation** For each generation setting we instantiate a language-specific prompt template with the sociodemographic attributes, issuing $n$=5 independent calls per model to capture stochastic variation. While annotating we consider the majority of annotations in the given 5 cases. The resulting English paragraphs form the source texts for the translation task. Every produced paragraph(generated or translated) is then labelled for honorific correctness by **GPT-4o**, following the automated annotation protocol described in Section 3.1. We sample and annotate $10\,000$ fictitious entities per language, yielding $40\,000$ annotated instances (4

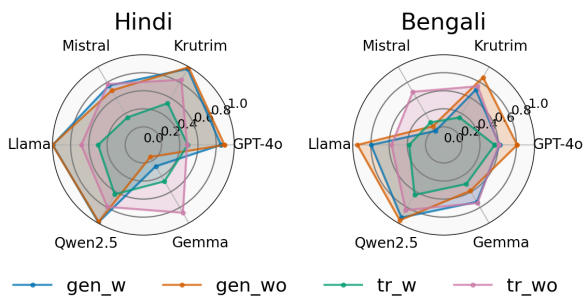

Figure 5: Comparison of honorific usage in LLMs for 4 different tasks – Gen_w, Gen_wo, Tr_w & Tr_wo, in Hindi and Bengali.

tasks × 2 languages × 10 000 examples).

## 4.2 Wiki vs LLM Comparison

The comparison between Wikipedia and LLM-generated outputs reveals nuanced differences in honorific usage across tasks and languages. While Wikipedia exhibits high honorific usage—86.6% in Hindi and 77.9% in Bengali, the LLMs vary widely in their alignment (see Figure 5). Interestingly, in Bengali, most LLMs (e.g., GPT-4o, Llama-3.1, Qwen-2.5, Krutrim-2) often exceed the Wikipedia baseline in generation tasks (especially Gen_wo), suggesting a possible overextension of politeness norms, whereas in Hindi, the LLMs frequently underperform relative to Wikipedia, particularly in translation tasks (e.g., GPT-4o's 56.1% in Tr_w vs. 86.6% in Wiki), indicating potential loss of socio-pragmatic cues. Among all models, Qwen-2.5 shows the highest consistency with Wikipedia, especially in Hindi. Notably, the translation-without-name (Tr_wo) setting shows erratic variation across both languages, highlighting that without clear identity cues, models struggle to replicate cultural norms consistently. Overall, while LLMs partially emulate Wikipedia patterns, their outputs are heavily modulated by language, task framing, and the presence of named entities, sometimes amplifying and sometimes suppressing expected cultural behaviors.

For feature-wise comparison of honorific usage between Wikipedia and LLM outputs, we present a detailed analysis using GPT-4o, as our Wikipedia annotations were also generated with GPT-4o, making it particularly insightful to examine the model's own preferences. Other LLMs also exhibit distinct honorific usage patterns that deviate from Wikipedia, similar to GPT-4o. Task-wise and language-wise results for all models are presented in Tables 6, 7, 8, and 9 in Appendix L.

When comparing the honorific usage patterns observed in Wikipedia with those generated by GPT-4o, several striking similarities and divergences emerge, varying both across languages (Hindi and Bengali) and prompting tasks (generation vs. translation, with and without names).

**Generation Tasks:** In general, GPT-4o's generation with name (Gen_w) exhibits closer alignment with Wikipedia patterns—particularly in Hindi. For instance, the prominence of entities like deities and famous individuals in eliciting honorifics is consistent across both Wiki and Gen_w settings. However, the generation without name (Gen_wo) setting reveals a tendency toward more neutral or flattened usage, where distinctions based on fame, age group, or gender are less sharply reflected than in Wikipedia.

**Translation Tasks:** Translation settings demonstrate stronger honorific polarization, especially in Hindi. GPT-4o translation with name (Tr_w) closely mirrors the culturally rich usage seen in Hindi Wikipedia, showing high sensitivity to features like deity status, fame, and age group. Interestingly, even in Bengali—where Wikipedia itself exhibits subtler honorific variation—the LLM in Tr_w appears more assertive, sometimes exaggerating cultural markings not strongly present in Wiki data. This over-alignment may stem from the model's exposure to culturally salient patterns during training. Conversely, translation without name (Tr_wo) settings show notable differences from both Wikipedia and other prompting variants. The absence of a named referent leads to reduced sensitivity to features like gender and fame. This effect is more pronounced in Bengali, possibly due to the language's lower baseline of honorific usage in Wikipedia, making the LLM default to less marked forms.

**Overall Similarities and Dissimilarities:** Across tasks, both languages show that GPT-4o generally respects some basic sociocultural norms, such as honorifics for deities or elders, but the consistency of the model with Wikipedia patterns is stronger when the prompt includes a name and is framed as a translation. Generation tasks, particularly without names, tend to dilute these norms, suggesting a context-dependent representation of cultural honorific practices. While Hindi aligns more robustly with Wikipedia across all tasks, Bengali shows divergence in subtle ways, highlighting how LLMs might learn stronger socio-pragmatic signals from languages with more overt honorific markers. Thus, GPT-4o is not merely replicating Wiki patterns but selectively amplifying or muting cultural cues based on input framing, **indicating that our initial hypothesis is wrong**.

## 5 Conclusion and Open Questions

Our study reveals several important insights: (i) we observe that Wikipedia articles often reflect inconsistent patterns in third-person pronoun and verb choices, (ii) LLMs exhibit their own internalized usage norms that diverge in nuanced ways from Wikipedia, hinting at the models' learned cultural inferences from large-scale web data, (iii) our socio-demographic feature-level comparisons indicate that factors like fame, age group, gender, and entity type significantly affect honorific usage patterns in both Wikipedia and LLMs, though the direction and degree of these effects vary. These findings offer an insight into how social and cultural norms surrounding respect and hierarchy are encoded, explicitly by human editors and implicitly by machine learning models. Yet, our work also opens up several critical avenues for future research:

**(1) Language-Aware Editorial Norms for Wikipedia.** Our analysis highlights inconsistencies in honorific usage across Wikipedia articles, suggesting the need for clearer, culturally sensitive editorial guidelines. A more language-aware framework could ensure both stylistic consistency and socio-cultural appropriateness, especially for languages with complex honorific systems.

**(2) Need for In-Depth Native Speaker Preference Studies.** While LLMs show distinct patterns and Wikipedia exhibits inconsistencies, our small-scale human study, where native experts were asked which honorific forms they would prefer when writing a Wikipedia-style article for a general online audience (see Details in Section L), revealed differences in overall honorific usages even between native speakers favoring for the same entities (see Figure 6). Feature-wise correlations of honorific usage also diverge from native expert judgments (see Figures 11 and 10 in Appendix L for detailed results). This calls for a more systematic and large-scale human study to understand the underlying personal and cultural factors shaping

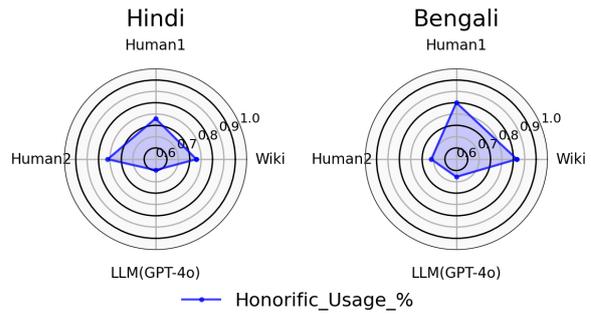

Figure 6: Comparison of honorific usage in LLM (GPT-4o-gen_w) vs Human Preferences vs Wikipedia.

honorific choices.

**(3) Honorifics in Second-Person and Casual Settings.** Our study focused exclusively on third-person usage in formal Wikipedia contexts. However, honorific dynamics can differ substantially in second-person usage and more casual or spoken communication. Future work should examine these alternate settings to provide a complete picture of honorific variation.

**(4) Cross-Linguistic and Regional Expansions.** Expanding this study to include other languages with rich honorific systems, such as Japanese, Korean, Javanese, or Tamil, could offer valuable comparative insights. Additionally, regional dialects and intra-language variation remain underexplored dimensions.

**(5) Impact of platform-specific editorial styles on honorific usage.** Wikipedia is governed by community norms, while news articles, fiction, or social media may follow different conventions. Investigating how honorifics vary across platforms can help delineate linguistic behavior driven by genre vs. cultural convention.

**(6) LLM Adaptation for Cultural Alignment.** Given that LLMs implicitly learn and propagate cultural norms, a crucial open question is how these models can be adapted or fine-tuned to better align with local cultural sensitivities, especially in sensitive applications such as education or public information systems. Our aim is to analyze a baseline model trained on our annotated data, which could provide a controlled point of comparison to understand how cultural norms are encoded.

Our work illustrates not only the power of computational methods in revealing socio-linguistic patterns but also the challenges that arise when these systems interact with deeply contextual, culture-bound linguistic phenomena like honorifics.

## Limitations

While our large-scale analysis of honorific usage in Hindi and Bengali Wikipedia articles offers valuable insights, several limitations should be noted.

First, the dataset is constrained to Wikipedia articles, which represent formal, encyclopedic language and may not capture more colloquial or informal uses of honorifics in everyday communication. This may limit the generalizability of our findings to other domains or registers of language.

Second, the use of *GPT-4o* for annotation, though validated with human expert evaluations, is not without its limitations. The LLM may struggle with nuanced or context-dependent honorific forms, especially in cases where the honorific usage is ambiguous or contextually fluid. While human validation mitigates this, some subtle sociolinguistic factors may still be missed by both the model and human annotators.

Third, the socio-socio-demographic features we examined, such as gender, age, and origin of culture, provide a useful framework for understanding honorific usage. However, this feature set is not exhaustive. Other cultural dimensions, such as regional dialects, social class, and profession, which could further influence honorific usage, were not included in this study.

Lastly, the reliance on static rules to extract and categorize Wikipedia articles may have introduced biases in the data selection process, potentially limiting the diversity of the extracted articles.

## Acknowledgments

This research was supported by the Microsoft Accelerate Foundation Models Research (AFMR) Grant. We thank all the team members involved in the internal pilot studies. Arora's lab is partly supported by grants from the Novo Nordisk Foundation (NNF24OC0099109), the Pioneer Centre for AI, and EU Horizon 2020 (101168951). We also gratefully acknowledge generous gifts from It-vest - networking universities.

# A  Language Selection Criteria

We began with the top 100 Wikipedia language editions by article count.[7] From this pool, we filtered 11 languages (*Japanese, Korean, Urdu, Tamil, Thai, Bengali, Hindi, Marathi, Javanese, Nepali, Bishnupriya Manipuri*) that theoretically support an obligatory distinction between honorific and non-honorific forms in third-person pronouns and verbs.[1] A preliminary investigation, involving 100 articles each from Japanese and Korean, revealed that Wikipedia articles in these languages predominantly use neutral pronouns and verbs, lacking explicit honorific distinctions. Thus, we excluded them from further analysis. Resource constraints further limited our focus to two languages, **Hindi** and **Bengali**, for which reliable native speaker expertise was available.

# B  Honorific Systems in Hindi & Bengali with Examples

Hindi (HI) and Bengali (BN), both major Indo-Aryan languages, exhibit rich and grammatically encoded honorific systems, particularly in second- and third-person references. Spoken by approximately 345 million (HI) and 237 million (BN) native speakers respectively in South Asia and their respective diasporas worldwide,[8] these languages obligatorily encode levels of social respect through pronominal and verbal morphology (Bhatt, 2012; Ray et al., 1966). Since Wikipedia articles are generally written in the third person, we focus our analysis on third-person honorific usage only. In Hindi, the third-person formal pronoun is वह (vah) paired with plural or subjunctive verb forms, whereas non-honorific references use either वह (vah) with singular verb forms or colloquial alternatives. In Bengali, formal third-person forms include তিনি (tini) with pluralized verb inflections, while non-honorifics use সে (shey) or ও (o) with singular verb forms. Verb agreement is key: honorific usage is often obligatorily marked through verb inflection even if the pronoun remains ambiguous (Bhatt, 2015; Douglas, 1985). Unlike nominal honorifics like the suffix "ji" in Hindi (or occasionally in Bengali), which are optional and stylistic, the pronominal-verb combinations are grammatically enforced in formal written contexts such as Wikipedia. Table 2 provides more concrete examples of honorific and non-honorific pronominal-verb pairs across both languages.

| Language | Pronouns Examples | Verbs Examples |
| --- | --- | --- |
| Hindi | *Honorific*: "वे" (they, plural), "उनका" (his/her, possessive), "उनके" (their, possessive), "उनको" (to them), "उन्होंने" (they did), "उन्हें" (to them), "उनसे" (from them), "इनका" (this person's), "इनके" (these people's), "इनको" (to these people)  *Non-Honorific*: "वह" (he/she), "वो" (they), "उसने" (he/she did), "उसको" (him/her), "उसे" (him/her), "उसका" (his/her, possessive), "उसके" (his/her, possessive) | *Honorific*: "उन्होंने किया" (they did), "उन्होंने कहा" (they said), "करते हैं" (do, plural), "मिलीं" (met, feminine plural), "थे" (were, plural)  *Non-Honorific*: "उसने किया" (he/she did), "उसने कहा" (he/she said), "करता है" (does, singular), "मिली" (met, feminine singular), "था" (was, singular) |
| Bengali | *Honorific*: "তিনি" (he/she), "তাঁরা" (they, plural), "তাঁর" (his/her, possessive), "তাঁদের" (their, possessive), "ইনি" (this person), "উনি" (he/she, informal), "ওঁনারা" (they, plural)  *Non-Honorific*: "সে" (he/she), "ও" (they), "ওরা" (they, informal), "তারা" (they, plural), "তার" (his/her, possessive), "ওর" (his/her, informal possessive), "তাদের" (their, possessive) | *Honorific*: "যান" (go), "বলেন" (says), "লেখেন" (writes), "করেছেন" (did), "বলেছেন" (said), "দেখেছেন" (saw), "শুনেছেন" (heard), "লিখেছেন" (wrote), "করবেন" (will do), "লিখবেন" (will write)  *Non-Honorific*: "যায়" (goes), "বলে" (says), "লেখে" (writes), "করেছে" (did), "বলেছে" (said), "দেখেছে" (saw), "শুনেছে" (heard), "লিখেছে" (wrote), "করবে" (will do), "লিখবে" (will write) |

Table 2: Examples of Third-Person Honorific and Non-Honorific Pronoun–Verb Constructions in Hindi and Bengali.

---

[7] https://meta.wikimedia.org/wiki/List_of_Wikipedias
[8] Hindi Wikipedia, Bengali Wikipedia

## C  Why Wikipedia as a Data Source

In selecting a corpus for studying third-person honorific usage, we carefully considered multiple alternatives, each with distinct advantages and limitations.

**Second-person honorifics.**  Although potentially insightful, second-person honorific usage is highly context-dependent, reflecting the immediate social relationship and power dynamics between interlocutors. Such dependencies make it unsuitable for establishing large-scale, generalizable baselines. We therefore restrict our study to third-person honorifics, which are more stable and consistently encoded in written text.

**Newspapers.**  News articles are a natural candidate for third-person references. However, our pilot study revealed that newspapers overwhelmingly employ honorifics across entities. This overuse likely stems from (a) sample bias, since newspapers predominantly cover celebrities, politicians, and other high-status individuals, and (b) editorial norms that enforce politically correct, respectful language. Consequently, newspaper corpora are poorly suited to uncovering fine-grained socio-demographic variation in honorific usage.

**Social media.**  Platforms such as X (formerly Twitter) and Facebook contain abundant third-person references, but these are often shaped by political stance, emotional states, and personal biases. For example, when a cricket team loses, fans may refer to players with non-honorifics as an expression of anger, despite ordinarily using honorifics for the same individuals. Such stance-driven variability is an interesting research question in its own right, but requires a controlled baseline for meaningful comparison.

**Wikipedia.**  In contrast, Wikipedia offers a middle ground between the formality of newspapers and the volatility of social media. It is a collaboratively curated knowledge resource where editorial guidelines encourage neutrality but allow community norms to surface in subtle linguistic choices. This makes Wikipedia particularly well-suited for establishing baseline distributions of honorific usage across socio-demographic categories. These distributions, in turn, provide a foundation for future studies on honorific usage in more contextually volatile settings, such as news or social media.

While Wikipedia provides an essential baseline for our present study, in ou future work, we aim to extend this work by systematically analyzing honorific usage across diverse sources such as newspapers, social media, and conversational data. This will allow us to compare how baseline community norms evolve or diverge in politically, emotionally, and socially situated contexts.

# D Dataset Statistics

| Feature | Hindi (Count/Percentage) | Bengali (Count/Percentage) |
|---|---|---|
| **Entity** | | |
| Human | 10585 (98.6%) | 10091 (95.6%) |
| Deity | 91 (0.8%) | 351 (3.3%) |
| Other Being | 48 (0.4%) | 94 (0.9%) |
| Animal | 7 (0.1%) | 18 (0.2%) |
| **Fame** | | |
| Famous | 10408 (97.0%) | 10119 (95.9%) |
| Controversial | 240 (2.2%) | 305 (2.9%) |
| Infamous | 83 (0.8%) | 130 (1.2%) |
| **Age Group** | | |
| Adult | 7553 (70.4%) | 7063 (66.9%) |
| Old | 2936 (27.4%) | 2796 (26.5%) |
| Not Applicable | 176 (1.6%) | 630 (6.0%) |
| Juvenile | 66 (0.6%) | 65 (0.6%) |
| **Gender** | | |
| Male | 7750 (72.2%) | 7630 (72.3%) |
| Female | 2941 (27.4%) | 2787 (26.4%) |
| Gender Neutral/Non-Specific | 40 (0.4%) | 137 (1.3%) |
| **Origin of Culture** | | |
| Native | 6665 (62.1%) | 5644 (53.5%) |
| Exotic | 4066 (37.9%) | 4910 (46.5%) |
| **Existence Type** | | |
| Real | 10425 (97.1%) | 9751 (92.4%) |
| Mythological | 301 (2.8%) | 687 (6.5%) |
| Fictional | 5 (0.0%) | 116 (1.1%) |
| **Life Status** | | |
| Alive | 6948 (64.7%) | 6237 (59.1%) |
| Dead | 3522 (32.8%) | 3578 (33.9%) |
| Not Applicable | 261 (2.4%) | 739 (7.0%) |
| **Era** | | |
| Modern | 10102 (94.1%) | 9367 (88.8%) |
| Historical | 629 (5.9%) | 1187 (11.2%) |
| **Pronoun/Verb in Wiki Article** | | |
| Honorific | 8360 (77.9%) | 9136 (86.6%) |
| Non-Honorific | 2371 (22.1%) | 1418 (13.4%) |
| **Total Count** | 10731 | 10554 |

Table 3: Wiki-dataset(Statistics) of Honorific usage in Hindi and Bengali using *GPT-4o*

# E Human Evaluation

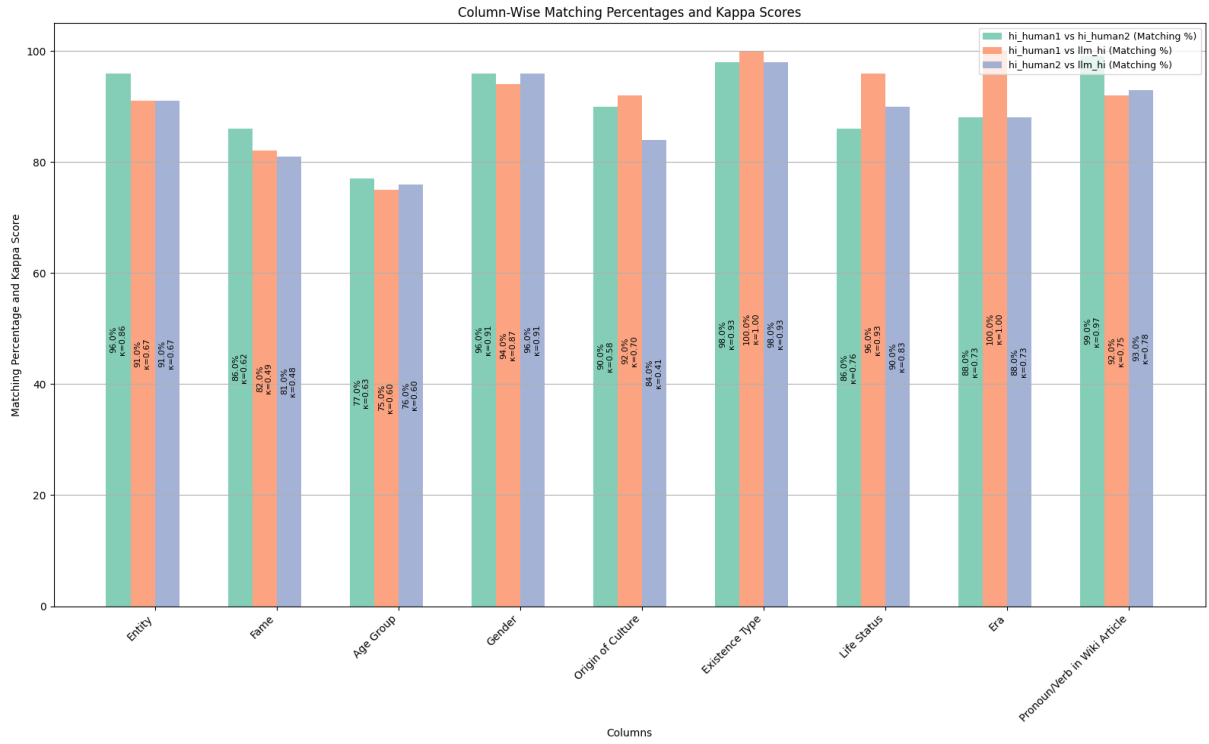

Figure 7: Human Evaluation for Hindi

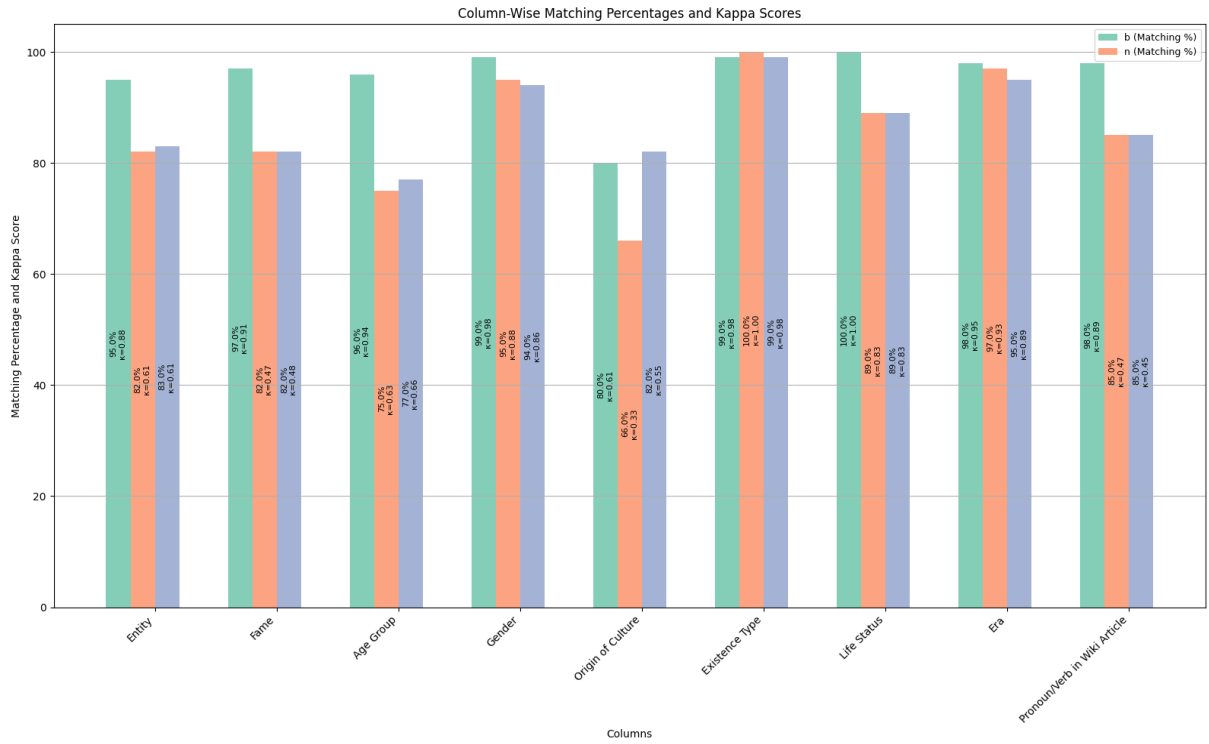

Figure 8: Human Evaluation for Bengali

|  | Hindi | | Bengali | |
| --- | --- | --- | --- | --- |
| Metric | Human1 vs LLM | Human2 vs LLM | Human1 vs LLM | Human2 vs LLM |
| Accuracy | 0.92 | 0.93 | 0.85 | 0.85 |
| Precision | 0.96 | 0.97 | 0.97 | 0.99 |
| Recall | 0.94 | 0.94 | 0.85 | 0.85 |
| F1 Score | 0.95 | 0.96 | 0.91 | 0.91 |

Table 4: Evaluation metrics summary for comparisons between human annotators and LLM (*GPT-4o*) for Hindi and Bengali.

# F Detailed Statistical Relevance Analysis of the Socio-cultural Features

**Chi-Square Test of Feature Association.** To evaluate whether socio-demographic features are statistically associated with honorific usage, we employed the Pearson chi-square test of independence. For each categorical feature (e.g., Gender, Fame, Origin of Culture), we constructed a contingency table crossing the feature's categories with the binary target variable (honorific vs. non-honorific usage). For instance, the Gender feature yields a $3 \times 2$ table: rows correspond to gender categories (male, female, neutral) and columns to counts of honorific and non-honorific usage.

Formally, the chi-square test statistic is defined as:

$$\chi^2 = \sum_{i=1}^{R} \sum_{j=1}^{C} \frac{(O_{ij} - E_{ij})^2}{E_{ij}}$$

where $O_{ij}$ and $E_{ij}$ denote the observed and expected frequencies in the cell at row $i$ and column $j$, and $R$ and $C$ are the number of rows and columns of the contingency table. The expected frequency $E_{ij}$ is computed under the null hypothesis of independence between the feature and honorific usage.

A significant chi-square value (with p-value $< 0.05$ after Bonferroni correction) indicates that the distribution of honorific vs. non-honorific usage differs meaningfully across the categories of that feature. In other words, knowing the feature provides non-trivial information about the likelihood of honorific usage. Importantly, the test does not specify the direction of association (e.g., which gender receives more honorifics); category-level directionality is examined separately through descriptive statistics and subsequent analyses.

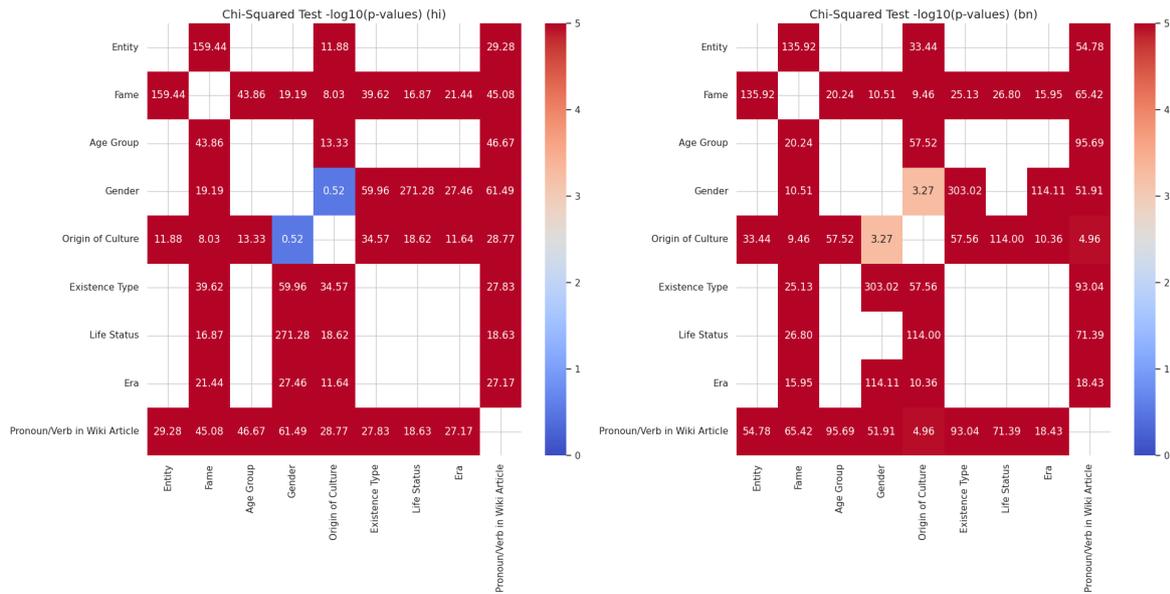

Figure 9: Chi-squared test heatmaps for Hindi (hi) and Bengali (BN) datasets showing the scores for various socio-demographic features associated with honorific and non-honorific usage in Wikipedia articles. Darker shades of red indicate higher statistical significance, while blue indicates lower significance. The most significant features across both datasets are Fame, Age Group, Gender, and Origin of Culture, as evidenced by the consistently high values across both languages. These features show the strongest associations with honorific usage, providing insights into how socio-cultural factors influence language patterns differently in Hindi and Bengali. The heatmaps highlight the cultural variability and the prominence of certain features in contributing to the distinction between honorific and non-honorific forms, thus allowing for a nuanced comparison of socio-cultural norms reflected in language use.

# G Annotation Guidelines for Humans

---

Annotation Guidelines for multilingual Honorifics Usage

## G.1 Overview

This study focuses on the (Bengali and Hindi) usage of honorific and non-honorific pronouns and verbs (verb-endings) in Wikipedia articles. Specifically, we analyze the third-person pronouns and verbs used to refer to individuals or entities, and annotate them as either honorific or non-honorific. This study examines linguistic choices in respect or formality, particularly within a socio-cultural context.

## G.2 Pronouns Overview

**Bengali Pronouns:** "সে", "তারা", "তাঁরা", "ও", "ওরা", "তিনি", "তাহারা", "তার", "তাঁর", "তাদের", "তাঁদের", "ওর", "ওদের", "তাহার", "তাহাদের"

**Hindi Pronouns:** "वह", "वे", "वो", "उसने", "उसको", "उसे", "उसका", "उसके", "उसकी", "उनका", "उनके", "उनकी", "उनको", "इसका", "इसकी", "इसके", "इनका", "इनकी", "इनके", "इनको", "उनसे", "उन्होंने", "उन्हें"

**Honorific Pronouns.** Used in formal or respectful contexts, often to show respect for the subject being referred to.

**Non-Honorific Pronouns.** Used in casual or informal contexts, typically when referring to peers, subordinates, or in less formal settings.

## G.3 Honorific and Non-Honorific Pronouns and Verbs

**Bengali Honorific Pronouns:** "তাঁরা", "তিনি", "তাঁর", "তাঁদের", "তাহার", "তাহাদের", "তারা" (Note: "তারা" and "তার" are used interchangeably with "তাঁরা" and "তাঁর")

**Bengali Non-Honorific Pronouns:** "সে", "তারা", "ও", "ওরা", "তাহারা", "তার", "তাদের", "ওর", "ওদের"

**Hindi Honorific Pronouns:** "वे", "उनका", "उनके", "उनकी", "उनको", "उन्होंने", "उन्हें", "उनसे", "इनका", "इनकी", "इनके", "इनको"

**Hindi Non-Honorific Pronouns:** "वह", "वो", "उसने", "उसको", "उसे", "उसका", "उसके", "उसकी", "इसका", "इसकी", "इसके"

## G.4 Examples of Honorific and Non-Honorific Verbs

**Bengali Honorific Verbs:** "করেছেন" (has done), "বলেছেন" (has said), "গেছেন" (has gone)
**Bengali Non-Honorific Verbs:** "করেছে" (has done), "বলেছে" (has said), "গেছে" (has gone)
**Hindi Honorific Verbs:** महात्मा गांधी ने देश की आज़ादी के लिए संघर्ष किया और स्वतंत्रता संग्राम के महानायक बने।
**Hindi Non-Honorific Verbs:** राजू बाहर खेलने चला गया।

## G.5 Annotation Instructions

- Open the provided Wikipedia article using the link.

- For each article, read the text and annotate the feature values and the type of pronouns (honorific or non-honorific) used based on the entity described.

- Use the following categories and guidelines to annotate the article:

# H LLM Probing Prompts

> **Prompts for different tasks**
>
> ## H.1 Generation Task with Name
>
> Write a detailed and vivid paragraph in $language$ only about $name$ who $present\_tense$ $fame$ $gender$ figure $era$ who $present\_tense$ a $role$. This $gender\_noun$ $present\_tense$ $age$ and $life\_status$. $capital\_subject$ $existence$ and $present\_tense$ $culture\_origin$ to the demography of the $language$-language speakers. Describe $possessive$ background, notable actions, and reputation, emphasizing $possessive$ role in $timeline$. Highlight the events that shaped $possessive$ legacy and discuss $possessive$ impact on society. Respond in $language$ only.
>
> ## H.2 Generation Task without Name
>
> Write a detailed and vivid paragraph in $language$ language about a completely fictional character with a made-up name, who $present\_tense$ $fame$ $gender$ figure $era$ and $present_tense$ a $role$. The story should be entirely imaginary, with no references to real people, places, or historical events. This $gender\_noun$ $present\_tense$ $age$ and $life\_status$. $capital\_subject$ $existence$ and $present\_tense$ $culture\_origin$ to the demography of the $language$-language speakers. Describe $possessive$ background, notable actions, and reputation, emphasizing $possessive$ role in $timeline$. Invent events that shaped possessive legacy and discuss possessive impact on society, ensuring all names and details are unique and fabricated.
>
> ## H.3 Translation with Name
>
> The following is a detailed and vivid paragraph in english about $name$. Translate the paragraph into $language$. Paragraph: $English\_Paragraph$
>
> ## H.4 Translation without Name
>
> The following is a detailed and vivid paragraph in $language$ about the entity in the paragraph. Translate the paragraph into $language$. Paragraph: $English\_Paragraph$
>
> ## H.5 Feature Categories for Annotation
>
> - **Entity:** God, Human, Animal, Other Being
>
> - **Fame (Sentiment associated with the entity):** Famous, Infamous, Controversial
>   *Famous:* Individuals known positively.
>   *Infamous:* Individuals known negatively.
>   *Controversial:* Entities who spark public debate due to their actions or opinions.
>
> - **Age Group:** Juvenile (under 18), Adult (18-60), Old (60+), N/A (for abstract entities such as gods or concepts)
>
> - **Gender:** Male, Female, Gender Neutral/Non-Specific
>
> - **Origin of Culture:** Native (from the same demographic where the language is spoken), Exotic (from a different demographic)
>
> - **Existence Type:** Real, Fictional, Mythological
>
> - **Life Status:** Alive, Dead, N/A (for abstract entities like gods or concepts)
>
> - **Era:** Historical (before 1800), Modern (after 1800)
>
> ## H.6 Pronoun and Verb Annotation in Articles
>
> You will also need to identify the type of pronouns/verbs used in the article:
>
> - **Pronouns/Verbs:** Honorific or Non-Honorific
>
> - Annotate based on the entity being described in the article. Select "Honorific" when the language is formal and respectful, and "Non-Honorific" for more casual or informal usage.
>
> **Note:** Use your judgment when selecting the best fit for each category. If you are unsure, refer to external sources or prior knowledge to assist in making an informed decision.

# I Prompt Details to annotate the honorific or non-honorific pronouns or verbs from Wiki Text

---

**Annotation Prompt: Honorific and Non-Honorific Usage in Bengali Wikipedia**

**Task Overview:** We need your help to specify the usage of honorific and non-honorific pronouns, and/or verbs in the Bengali Wikipedia article titled "দুর্গা".

**Examples of Pronouns:**

- Honorific Pronouns: "তাঁরা", "তিনি", "তাঁর", "তাঁদের"
- Non-Honorific Pronouns: "সে", "তারা", "ও", "ওদের"

**Examples of Verbs:**

- Honorific Verbs: "করেছেন", "বলেছেন"
- Non-Honorific Verbs: "করেছে", "বলেছে"

**Article Text (excerpt):** দুর্গা (; অর্থাৎ "যিনি দুর্গতি বা সংকট থেকে রক্ষা করেন"; এবং "যে দেবী দুর্গম নামক অসুরকে বধ করেছিলেন")...তাঁকে আদ্যাশক্তির রণরঙ্গিনী এক মহাদেবীর রূপ বলে মান্য করেন।
**Annotation Instructions:** Carefully review the article and select the correct values for the following features:
**Features:**

- **Entity:** God, Human, Animal, or Other Being
- **Fame (Sentiment):** Famous, Infamous, Controversial
- **Age Group:** Juvenile (under 18), Adult (18-60), Old (60+), Not Applicable
- **Gender:** Male, Female, Gender Neutral/Non-Specific
- **Role:** Politics, Science, Arts, Entertainment, Religion, Sports, Business, etc.
- **Origin of Culture:** Native, Exotic
- **Existence Type:** Real, Fictional, Mythological
- **Life Status:** Alive, Dead, Not Applicable
- **Era:** Historical (before 1800), Modern (after 1800)
- **Pronoun/Verb in Wiki Article:** Honorific, Non-Honorific

**Additional Considerations:**

- **Pronoun/Verb in a Written Setting:** If writing about this entity in an article or blog, which pronouns/verbs would you use?
- **Pronoun/Verb in a Spoken Setting:** If discussing the entity with friends, which pronouns/verbs would you use?

**Output Format:** ONLY provide your answers in dictionary format:

```
{
    'Entity': 'God',
    'Fame': 'Famous',
    'Age Group': 'N/A',
    'Gender': 'Female',
    'Role': 'Deity',
    'Origin of Culture': 'Native',
    'Existence Type': 'Mythological',
    'Life Status': 'N/A',
    'Era': 'Historical',
    'Pronoun/Verb in Wiki Article': 'Honorific',
    'Pronoun/Verb in Written Setting': 'Honorific',
    'Pronoun/Verb in Spoken Setting': 'Non-Honorific'
}
```

## J  Clarification on the Role of GPT-4o in Annotation and Evaluation

A potential concern is whether our use of GPT-4o for both annotation and evaluation introduces circularity or bias. We address this explicitly here.

**Annotation Task.**  GPT-4o was employed as a classifier to annotate honorific vs. non-honorific usage in Wikipedia and LLM-generated text. This task focused on explicit grammatical markers (e.g., third-person pronouns and verb inflections), where GPT-4o demonstrated high accuracy due to its linguistic precision. These annotations were further validated by human experts, achieving substantial agreement. Importantly, in this role GPT-4o did not judge the *appropriateness* of honorific use but merely identified which form appeared in the text.

**Probing Task.**  In contrast, LLM evaluation involved probing models, including GPT-4o itself, through controlled generation and translation tasks with distinct prompts. Here, the objective was to study model preferences in producing honorific or non-honorific forms under different cultural and contextual conditions.

**Separation of Roles.**  These two uses of GPT-4o are methodologically independent: the annotation task identifies what form was used, while the probing task elicits what form a model chooses to generate. Even if GPT-4o produced an "incorrect" honorific form in the probing stage, the annotation stage would simply mark the form present, without privileging GPT-4o's output or evaluating its correctness. Thus, there is no circularity in our design.

In summary, GPT-4o served in two distinct capacities: (i) as a high-precision annotator of surface linguistic forms and (ii) as one of several models under evaluation for generative behavior. Their separation ensures that our findings are not confounded by overlap between annotation and evaluation.

## K  Annotators Demographic

The annotation process for this study involved four native speakers, two of Hindi and two of Bengali. The annotators were selected to ensure a diverse mix of gender and academic background, providing both linguistic expertise and native-level language proficiency.

**Hindi Annotators:** The two Hindi annotators are native speakers. Both have a strong academic background in linguistics. Their extensive knowledge of the Hindi language and its socio-cultural nuances allowed for accurate annotation of honorific and non-honorific forms.

**Bengali Annotators:** Similarly, the Bengali annotators are native speakers. They are computational linguistics. Their native expertise in Bengali, coupled with their research experience, ensured high-quality and culturally accurate annotations.

This mix of linguistic researchers, at both the Master's and Ph.D. levels, male and female, provided a balanced and informed perspective during the annotation process, ensuring that cultural and linguistic subtleties in both languages were accurately captured.

## L  Human Preferences and Socio-Demographic dependencies

**Eliciting Native Usage Preferences.**  We instructed the annotators to indicate their preferred usage of honorifics or non-honorifics for the same entity in a written context similar to that of Wikipedia (on a randomly selected sample of 100 entries from a total of 10,000 Wikipedia articles). The goal was to analyze the differences in usage preferences between the annotators' choices and the conventions typically found in Wikipedia-style writing. To guide their decision, the annotators were given the following instruction: 'If you were writing an article (Wikipedia-style article or blog post intended for a general online audience) about the same entity to be published on the web for a general audience, what kind of pronouns/verbs would you use? (Honorific/Non-Honorific).' This instruction aimed to help them choose their preferred usage for each entity. We used these preferences to compare against the honorific usage patterns identified in the Wikipedia articles, which are discussed in the following section. After the annotation process, we computed the inter-annotator agreement scores, which were 0.75 for Hindi and 0.60 for Bengali annotators, respectively.

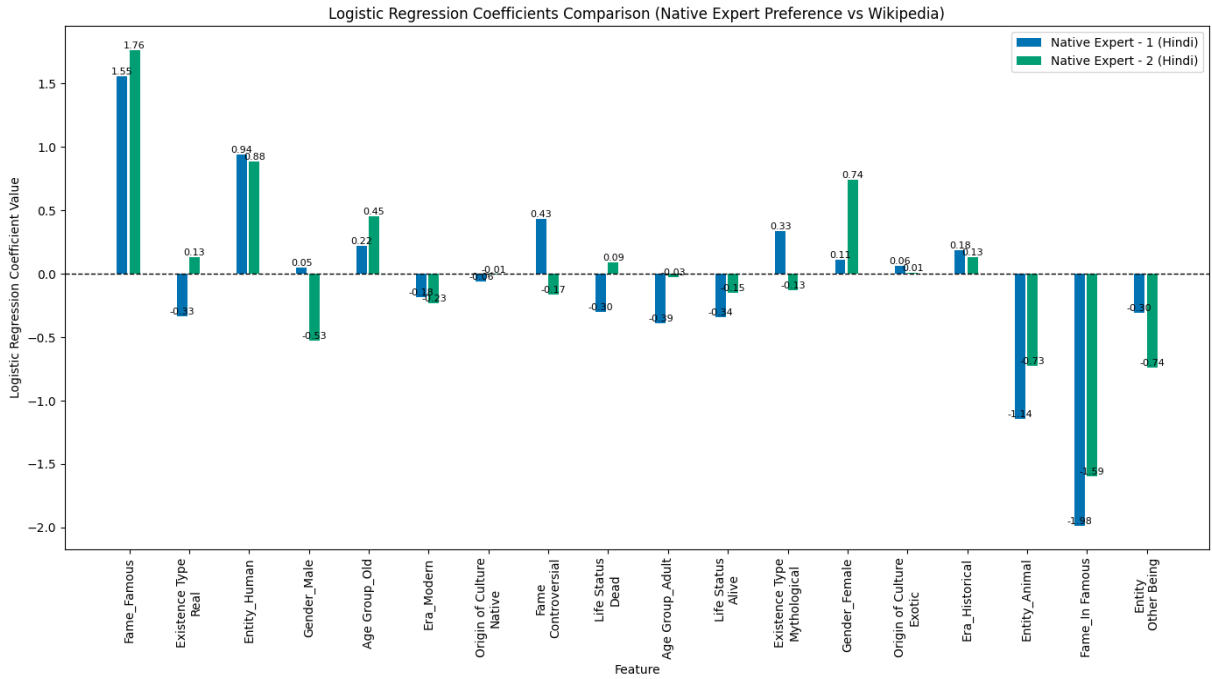

Figure 10: Comparing the logistic regression coefficients human preferences in hindi.

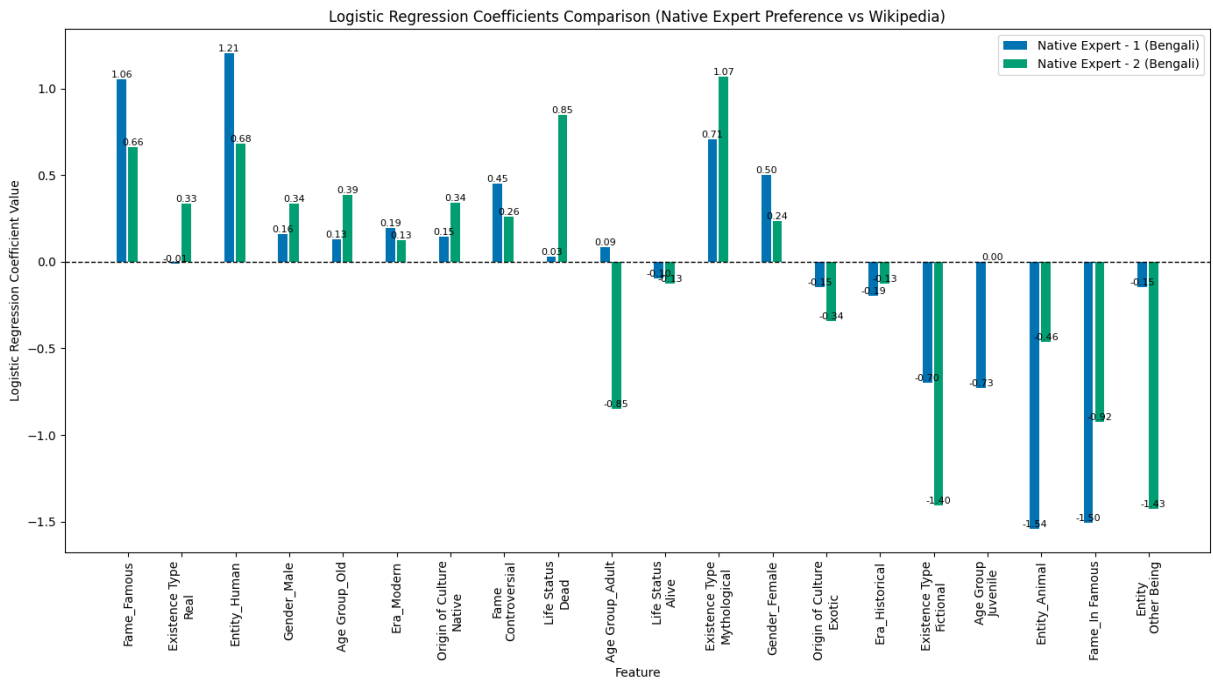

Figure 11: Comparing the logistic regression coefficients human preferences in hindi.

## M  Wiki Analysis for All Socio-Demographic Features

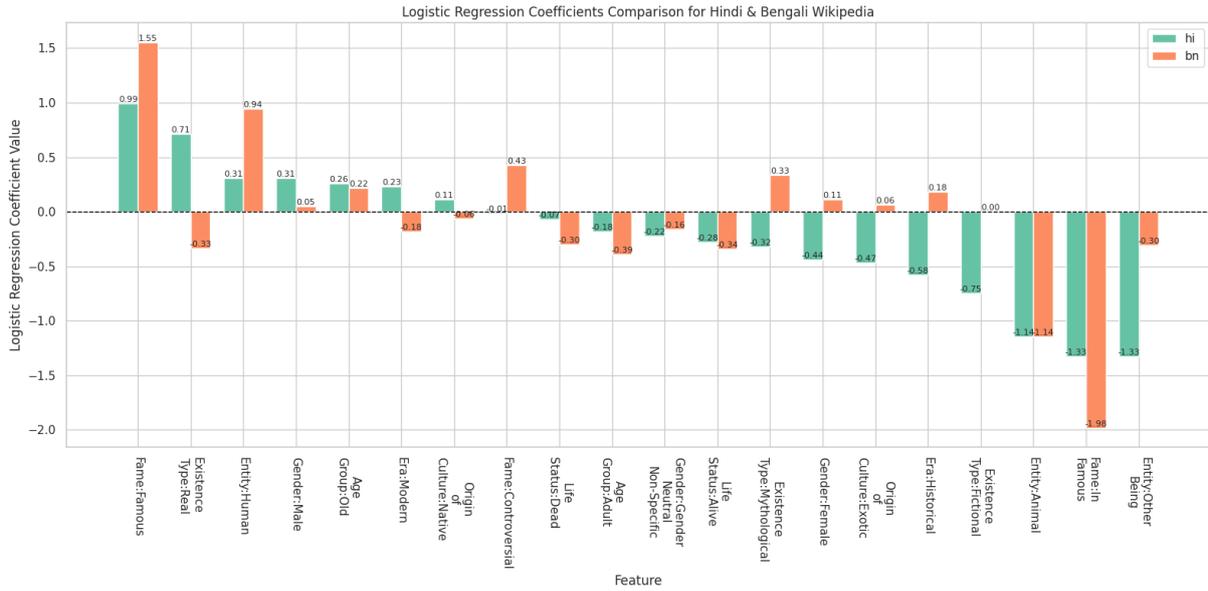

Figure 12: Comparing the logistic regression coefficients for 10,000 randomly sampled and annotated pages for Hindi and Bengali Wikipedia, with the x-axis showing socio-demographic factors affecting honorific usage.

## N  Honorific Usage Patterns in LLMs

| Scenario | Hindi | | | | Bengali | | | |
|---|---|---|---|---|---|---|---|---|
| | GEN_W | GEN_WO | TR_W | TR_WO | GEN_W | GEN_WO | TR_W | TR_WO |
| GPT-4o | 61.2 | 80.8 | 56.1 | 60.8 | 85.3 | 89.1 | 48.8 | 48 |
| Llama-3.1 | 79.8 | 94.7 | 37.9 | 57.2 | 99.6 | 100 | 49.8 | 68.3 |
| Gemma | 72.9 | 58.7 | 49.9 | 74.3 | 27.3 | 15.4 | 46.5 | 86.8 |
| Mistral | 17.9 | 23.9 | 28.8 | 67.9 | 75.8 | 69.6 | 35.2 | 77.8 |
| Qwen-2.5 | 92.9 | 96.4 | 64 | 83.5 | 99 | 98.3 | 63.2 | 79.2 |
| Krutrim-2 | 70.6 | 86.5 | 35.1 | 75 | 97.3 | 99.4 | 53.3 | 83.3 |
| *Wikipedia Usage* | | 77.9 | | | | 86.6 | | |
| *Human Preferences* | | 81 | | | | 91 | | |

Table 5: Honorific-usage percent across different scenarios from LLMs to Wikipedia across four probe settings. "Gen" = generation task, "Tr" = translation task; "w" = name included, "wo" = no name. Wikipedia Usage shows the usage of honorifics for introduction in hindi and bengali in wikipedia and Human Preferences shows the human annotator preferences in honorific usage for these languages.

Table 6: Honorific–usage percentages for eleven socio-demographic feature buckets per language for GEN_W. **EH**: Entity_Human, **EA**: Entity_Animal, **GM**: Gender_Male, **GF**: Gender_Female, **AA**: Age_Adult, **AJ**: Age_Juvenile, **AO**: Age_Old, **OC$_N$**: Origin_Native, **OC$_E$**: Origin_Exotic, **FF**: Fame_Famous, **FI**: Fame_Infamous.

| Model | Hindi | | | | | | | | | | | Bengali | | | | | | | | | | |
|---|---|---|---|---|---|---|---|---|---|---|---|---|---|---|---|---|---|---|---|---|---|---|
| | EH | EA | GM | GF | AA | AJ | AO | OC$_N$ | OC$_E$ | FI | FF | EH | EA | GM | GF | AA | AJ | AO | OC$_N$ | OC$_E$ | FI | FF |
| GPT-4o | 0.27 | -1.05 | -0.31 | -0.07 | 0.03 | -0.46 | 0.17 | -0.44 | 0.44 | -1.11 | 1.65 | 0.43 | -0.96 | -0.18 | 0.50 | 0.02 | -1.46 | 0.64 | -0.01 | 0.01 | -0.46 | 1.22 |
| Gemma | 0.47 | -2.02 | -0.20 | -0.81 | 0.67 | -0.61 | 0.41 | 0.33 | -0.32 | -0.70 | 1.43 | -1.09 | 0.13 | 0.23 | -0.17 | 0.27 | -0.81 | 0.10 | 0.18 | -0.19 | -0.30 | 0.50 |
| Krutrim | 0.32 | -1.76 | -0.76 | 1.34 | -0.78 | 0.60 | 0.34 | 0.09 | -0.10 | -2.28 | 1.47 | -0.25 | -1.37 | 0.32 | 0.74 | 0.13 | 0.79 | -0.14 | 0.43 | -0.41 | -0.64 | 1.22 |
| Llama | -0.24 | -0.98 | -0.66 | -0.27 | -0.06 | 0.37 | -0.00 | 0.53 | -0.53 | -0.38 | 0.96 | 0.20 | -1.22 | -0.19 | 0.57 | -0.39 | 0.21 | 0.21 | -0.62 | 0.61 | 0.56 | -0.31 |
| Mistral | 0.24 | 0.15 | -0.35 | -0.58 | 0.15 | -0.03 | -0.05 | 0.27 | -0.28 | -0.65 | 0.70 | -0.28 | -0.93 | 0.16 | -0.22 | 0.05 | -0.26 | -0.22 | 0.17 | -0.14 | -0.59 | 0.68 |
| Qwen2.5 | -0.04 | -1.57 | 0.43 | -0.96 | 0.40 | -0.96 | 0.91 | 0.10 | -0.10 | 0.62 | 0.66 | 0.39 | 0.04 | 0.04 | -0.69 | 0.09 | 0.04 | 0.01 | 0.97 | -0.98 | -0.19 | 0.06 |

Table 7: Honorific–usage percentages for eleven socio-demographic feature buckets per language for GEN_WO. **EH**: Entity_Human, **EA**: Entity_Animal, **GM**: Gender_Male, **GF**: Gender_Female, **AA**: Age_Adult, **AJ**: Age_Juvenile, **AO**: Age_Old, **OC$_N$**: Origin_Native, **OC$_E$**: Origin_Exotic, **FF**: Fame_Famous, **FI**: Fame_Infamous.

| Model | Hindi | | | | | | | | | | | Bengali | | | | | | | | | | |
|---|---|---|---|---|---|---|---|---|---|---|---|---|---|---|---|---|---|---|---|---|---|---|
| | EH | EA | GM | GF | AA | AJ | AO | OC$_N$ | OC$_E$ | FI | FF | EH | EA | GM | GF | AA | AJ | AO | OC$_N$ | OC$_E$ | FI | FF |
| GPT-4o | 0.53 | -1.03 | 0.17 | 0.14 | 0.46 | -1.06 | 0.81 | 0.57 | -0.56 | -1.41 | 0.82 | 0.80 | -2.78 | -0.25 | 0.03 | 0.04 | -1.12 | 0.63 | 0.07 | -0.03 | -1.24 | 0.33 |
| Gemma | -0.44 | -1.19 | 0.15 | 0.01 | 0.31 | -0.44 | 0.92 | 0.12 | -0.05 | 0.23 | 0.89 | -0.14 | -0.00 | -0.10 | 0.05 | 0.40 | -0.49 | -0.00 | 0.66 | -0.67 | 0.27 | -0.17 |
| Krutrim | 0.79 | -1.97 | 0.11 | -0.22 | 0.63 | -1.42 | 1.30 | 0.74 | -0.74 | -0.14 | 0.95 | -0.30 | 0.01 | -0.86 | 0.60 | -0.18 | -0.26 | 0.07 | -0.39 | 0.39 | -0.80 | 0.19 |
| Llama | - | - | - | - | - | - | - | - | - | - | - | - | - | - | - | - | - | - | - | - | - | - |
| Mistral | -1.31 | 0.64 | -0.19 | 0.32 | 0.20 | 0.28 | -0.22 | 0.79 | -0.80 | 0.48 | 0.61 | -0.49 | -0.75 | 0.10 | -0.30 | 0.08 | 0.81 | -0.26 | -0.13 | 0.17 | -0.35 | 0.03 |
| Qwen2.5 | 0.05 | 0.03 | -0.20 | -0.33 | -0.84 | 0.32 | 0.46 | -0.01 | 0.02 | 0.19 | 0.27 | -0.28 | 0.02 | -0.09 | 0.34 | 0.10 | -1.22 | 0.72 | 0.34 | -0.33 | -0.17 | -0.11 |

Table 8: Honorific–usage percentages for eleven socio-demographic feature buckets per language for TR_W. **EH**: Entity_Human, **EA**: Entity_Animal, **GM**: Gender_Male, **GF**: Gender_Female, **AA**: Age_Adult, **AJ**: Age_Juvenile, **AO**: Age_Old, **OC$_N$**: Origin_Native, **OC$_E$**: Origin_Exotic, **FF**: Fame_Famous, **FI**: Fame_Infamous.

| Model | Hindi | | | | | | | | | | | Bengali | | | | | | | | | | |
|---|---|---|---|---|---|---|---|---|---|---|---|---|---|---|---|---|---|---|---|---|---|---|
| | EH | EA | GM | GF | AA | AJ | AO | OC$_N$ | OC$_E$ | FI | FF | EH | EA | GM | GF | AA | AJ | AO | OC$_N$ | OC$_E$ | FI | FF |
| GPT-4o | 1.35 | -0.17 | -0.40 | -0.42 | 0.21 | -0.95 | 0.05 | -0.15 | 0.15 | -0.95 | 2.24 | -0.34 | -0.68 | 0.15 | 0.35 | 0.55 | -1.54 | 0.64 | 0.01 | -0.04 | -2.19 | 2.30 |
| Gemma | 0.56 | -0.38 | -0.42 | 0.30 | -0.24 | -0.87 | 0.65 | -0.59 | 0.61 | -0.43 | 1.42 | 1.52 | -1.19 | 0.16 | 0.30 | 0.28 | -1.48 | 0.37 | -0.50 | 0.48 | 0.13 | 1.47 |
| Krutrim | 0.46 | -1.11 | -0.24 | 0.46 | 0.13 | -1.15 | 0.65 | -0.17 | 0.17 | -1.70 | 2.63 | 0.78 | -1.15 | -0.19 | 0.09 | 0.01 | -1.02 | -0.35 | -0.21 | 0.20 | -1.50 | 2.55 |
| Llama | 0.31 | -0.14 | -0.47 | 0.04 | -0.15 | -0.87 | 0.84 | -0.54 | 0.51 | -0.92 | 1.96 | 0.49 | -1.29 | -0.13 | 0.46 | -0.43 | -0.88 | 0.76 | -0.24 | 0.22 | -0.47 | 1.57 |
| Mistral | 0.66 | -1.28 | 0.17 | -0.52 | 0.42 | -0.28 | 1.00 | -0.32 | 0.30 | -0.44 | 1.85 | 1.86 | -1.00 | -0.00 | 0.21 | 0.24 | -1.01 | 0.28 | -0.43 | 0.43 | -0.19 | 0.71 |
| Qwen2.5 | 1.57 | -2.71 | -0.33 | -1.43 | -0.48 | -0.52 | 0.58 | 0.13 | -0.08 | -0.67 | 1.21 | 0.92 | -1.72 | 0.15 | -0.19 | -0.16 | -0.91 | -0.19 | 0.07 | -0.08 | -0.55 | 1.46 |

Table 9: Honorific–usage percentages for eleven socio-demographic feature buckets per language for TR_WO. **EH**: Entity_Human, **EA**: Entity_Animal, **GM**: Gender_Male, **GF**: Gender_Female, **AA**: Age_Adult, **AJ**: Age_Juvenile, **AO**: Age_Old, **OC$_N$**: Origin_Native, **OC$_E$**: Origin_Exotic, **FF**: Fame_Famous, **FI**: Fame_Infamous.

| Model | Hindi | | | | | | | | | | | Bengali | | | | | | | | | | |
|---|---|---|---|---|---|---|---|---|---|---|---|---|---|---|---|---|---|---|---|---|---|---|
| | EH | EA | GM | GF | AA | AJ | AO | OC$_N$ | OC$_E$ | FI | FF | EH | EA | GM | GF | AA | AJ | AO | OC$_N$ | OC$_E$ | FI | FF |
| GPT-4o | 0.71 | 0.31 | 0.25 | 0.16 | 0.19 | -1.78 | 1.06 | -0.02 | 0.06 | -0.72 | 1.80 | 0.49 | -0.81 | 0.03 | 0.10 | -0.22 | -0.66 | 0.98 | -0.10 | 0.02 | -0.53 | 1.21 |
| Gemma | 0.06 | -1.92 | 0.36 | 0.66 | -0.10 | -0.99 | 0.19 | -0.64 | 0.64 | -0.14 | 1.58 | 0.70 | -1.02 | -0.09 | 0.07 | -0.05 | -0.90 | 0.87 | -0.63 | 0.64 | -0.64 | 0.77 |
| Krutrim | 0.67 | 0.06 | -0.29 | 0.10 | 0.25 | -0.91 | 0.60 | -0.34 | 0.34 | -0.80 | 2.32 | -0.47 | -0.09 | -0.32 | -0.65 | -0.21 | -0.40 | -0.16 | -0.34 | 0.35 | -0.64 | 1.99 |
| Llama | -0.78 | -0.68 | 0.30 | 0.28 | 0.02 | -1.17 | 1.29 | -1.74 | 1.76 | -0.38 | 2.17 | 0.35 | -0.09 | -0.39 | -0.16 | -0.22 | -1.18 | 1.26 | -1.30 | 1.31 | -0.56 | 2.16 |
| Mistral | 0.25 | -0.55 | -0.01 | 0.01 | -0.05 | 0.04 | -0.04 | -1.60 | 1.61 | -0.21 | 1.12 | 0.01 | -0.97 | -0.15 | 0.09 | 0.03 | 0.07 | 0.63 | 0.03 | -0.02 | 0.05 | 0.58 |
| Qwen2.5 | 0.53 | -1.54 | -0.38 | 0.13 | 0.24 | -0.93 | 0.77 | -1.17 | 1.17 | -0.22 | 1.36 | 0.11 | -0.47 | -0.10 | 0.06 | -0.37 | -0.23 | -0.08 | -1.10 | 1.11 | -0.49 | 0.95 |